# Sequential Point Cloud Prediction in Interactive Scenarios: A Survey

Haowen Wang
School of Mechanical Engineering
Beijing Institute of Technology
Beijing, China
wang.h.w@outlook.com

Zirui Li
School of Mechanical Engineering
Beijing Institute of Technology
Beijing, China
3120195255@bit.edu.cn

Jianwei Gong
School of Mechanical Engineering
Beijing Institute of Technology
Beijing, China
gongjianwei@bit.edu.cn

*Abstract*—Point cloud has been widely used in the field of autonomous driving since it can provide a more comprehensive three-dimensional representation of the environment than 2D images. Point-wise prediction based on point cloud sequence (PCS) is an essential part of environment understanding, which can assist in the decision-making and motion-planning of autonomous vehicles. However, PCS prediction has not been deeply researched in the literature. This paper proposes a brief review of the sequential point cloud prediction methods, focusing on interactive scenarios. Firstly, we define the PCS prediction problem and introduce commonly-used frameworks. Secondly, by reviewing non-predictive problems, we analyze and summarize the spatio-temporal feature extraction methods based on PCS. On this basis, we review two types of PCS prediction tasks, scene flow estimation (SFE) and point cloud location prediction (PCLP), highlighting their connections and differences. Finally, we discuss some opening issues and point out some potential research directions.

## I. Introduction

Perceiving the surrounding environment is vital to autonomous driving [1, 2]. A comprehensive understanding of the environment is the cornerstone of specific tasks such as route planning [3-6] and driving decision-making [7-9]. In contrast to two-dimensional images, which suffer from light and occlusion problems, point clouds retain the complete spatial structure of the three-dimensional space, providing rich geometry and scale information [10]. Recently, learning-based methods have shown their application prospects in processing point clouds. In particular, the pioneering work PointNet [11] and PointNet++ [12] make it possible to process 3D point clouds directly without projection or voxelization. The structural features of the 3D space are therefore fully utilized globally and locally.

Most of the existing works focus on the acquisition and understanding of spatial features in a single scene, i.e., only one point cloud frame is input to the model for training at a time. Autonomous driving vehicles should also specify corresponding execution strategies based on the movement of objects in the changing environment [13]. However, forecasting the future in a dynamic scene is still one of the urgent problems to be solved. Point cloud sequence (PCS) contains spatio-temporal information of scenes obtained by arranging LiDAR point cloud frames in time series. Based on PCS, some studies aim to capture the transfer of states in the time dimension and combine them with spatial features to achieve better results [14-16].

After encoding the spatio-temporal features between scenes, one can make predictions at a high-level or low-level, depending on the research object. High-level predictions focus on entities segmented from point clouds. After tracking the historical trajectory of the objects through target recognition or semantic segmentation, trajectory prediction method can be used to predict the movement of the instance [17-21]. However, two main problems exist: 1) Lack of background information, which may contain potential factors affecting the prediction result. Moreover, the context can associate entities in the environment to reflect the interaction [22]. 2) Low processing efficiency. When there are multiple entities around the host vehicle, each of them needs to be detected and predicted separately. This isolated prediction is not only inefficient but also unable to encode the interactive behavior between instances.

In the real driving environment, each vehicle may have a potential impact on other vehicles around it, so it is necessary to model the interaction between vehicles. The interaction scene contains the interaction behavior of vehicles. Compared with the general point cloud scenes which describe a single object, the number of scanning points is much more in interactive scene, the feature information behind the scanning points is more complex, and each scene may contain multiple target objects.

Part of recent research predicts point cloud at a low level [22-25], that is, by learning the temporal and spatial relationship of points in historical observations, predicting the movement changes of points in the entire scene at a future time. This point-wise prediction implies the prediction of abstract instances. However, it can better characterize the interaction behavior since it has better feature extraction ability, avoiding the above problems. Related research is still in the early stage.

Given the importance of environmental understanding in autonomous driving and the huge development potential of low-level point-cloud prediction, this article systematically reviews the current point cloud prediction methods, with particular attention to the point-wise point cloud prediction in interactive scenarios, which is different from current point cloud review papers such as [26, 27]. The related methods usually include hierarchical feature extraction to achieve different scale feature abstraction, and can deal with large point cloud scenes to adapt to the interactive environment of autonomous vehicles. We first summary the spatio-temporal feature modeling methods of PCS and then analyze the low-level point cloud prediction methods in detail. As far as we know, this is the first review related to the sequential point-cloud prediction problem in interactive conditions.

This article is organized as follows. Section II defines the point cloud prediction problem and introduces the commonly used framework. Section III reviews the spatiotemporal feature modeling methods in non-prediction problems. Section IV summarizes the current PCS prediction methods according to the different prediction forms, emphasizing the differences and connections between methods. Section V discusses the remaining problems in point cloud prediction and prospects for future research directions.

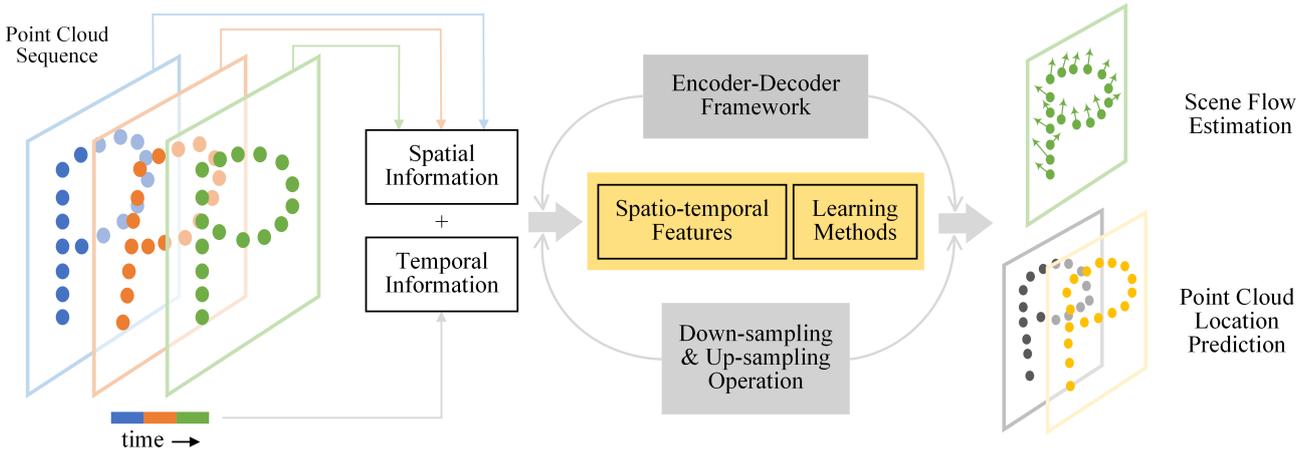

Figure 1. General framework of PCS prediction problem

## II. PROBLEM DEFINITION AND RELATED FRAMEWORKS

This section contains three parts. We first define the sequential point cloud prediction problem and clarifies its input and output. In the next part, we explain the commonly used "encoder-decoder" framework in sequence-to-sequence (Seq2Seq) problem. Finally, an efficient operation called "down-sampling & up-sampling" is introduced, which is helpful in dealing with large-scale interactive scenes. The general framework for PCS prediction is shown in Figure 1. Details will be explained in the following sections.

### A. Problem Definition

Point Cloud Sequence (PCS) is an ordered collection of point cloud frames with rich temporal information. A three-dimensional PCS with a length of $T$ can be described as $S = (S_1, S_2, ..., S_t, ..., S_T)$, where $S_t$ is the point cloud frame at the time $t$. We denote a point cloud frame with $N_t$ points by $S_t = \{(\mathbf{x}_i^{(t)}, \mathbf{f}_i^{(t)}) | i = 1, 2, ..., N_t\}$, where $\mathbf{x}_i^{(t)} \in \mathbb{R}^3$ is the coordinate vector and $\mathbf{f}_i^{(t)} \in \mathbb{R}^c$ is the feature vector. Frames are arranged in an orderly manner from the perspective of PCS, but points are disordered in each point cloud frame. PCS can also be regarded as 4D point cloud data.

The prediction of the sequential point cloud can be treated as a Seq2Seq problem. Specifically, given a historical PCS $S_{obs} = (S_{-M+1}, S_{-M+2}, ..., S_0)$ of length $M \in \mathbb{N}^+$, the model should output a predicted PCS $S_{pred} = (S_1, S_2, ..., S_N)$ of length $N \in \mathbb{N}^+$. We generally require $M > 1, N \geq 1$ to predict the point cloud at one or more time steps by learning the temporal and spatial characteristics of the PCS over a period.

### B. The "Encoder-Decoder" Framework

The "encoder-decoder" framework is widely used in the sequential data processing, as shown in Figure 2. Since it is difficult to preset the driving behavior, we need to express the vehicle interaction features in the scene through efficient feature extraction method in an end-to-end fashion. In our problem, the "encoder" can be regarded as the feature abstraction of complex and irregular point cloud data, extracting and integrating the spatio-temporal features into a hidden vector form which is more conducive to learn. In comparison, the "decoder" maps the updated hidden features into Euclidean space to achieve PCS prediction.

As a concrete realization of the Seq2Seq problem, the "encoder-decoder" framework has many advantages in PCS prediction problems, which can be summarized in the following three aspects: regular processing of complex point cloud structures; encoded spatio-temporal features are easy to transfer between scenes; well-matched with classic model architectures. In the non-prediction problem with PCS, the characterization of spatiotemporal features can also be regarded as an encoding process. However, the result of decoding will no longer reflect motion.

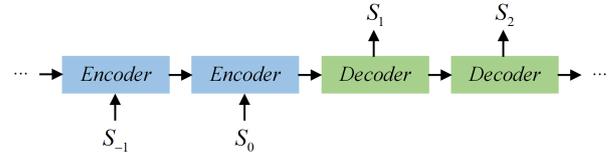

Figure 2. "Encoder-Decoder" Framework

### C. The "Down-sampling & Up-sampling" Operation

For safety reasons, autonomous vehicles need to make high-speed and high-precision predictions of environmental changes to support decision-making and motion-planning in real-time, such as avoiding collisions with moving pedestrians [8]. However, the massive point cloud in the interactive scene brings a heavy burden to the calculation. PointNet++ [12] proposes a hierarchical feature extraction method, which uses the "sampling & grouping" operation to down-sample the points in the scene and up-sample the spatial features to restore the entire scene by feature propagation layer. While reducing the number of points processing in the intermediate layers, the above operation realizes the extraction and integration of spatial features in different scales, reflecting high application value in processing large-scale point cloud data.

The "down-sampling & up-sampling" operation represented by [12] can extract local features of different scales, which is conducive to focus on and integrate relatively small instance areas and relatively large interaction areas, and can therefore modeling interactive behavior. Moreover, with the reduction of computation, the model can deal with more scanning points under the same hardware conditions.

## III. SPATIO-TEMPORAL RELATIONSHIP MODELING FOR NON-PREDICTIVE PROBLEMS

The low-level prediction of PCS relies on a deep understanding of the spatio-temporal features between scenes. Therefore, the extraction and transmission of spatio-temporal features are essential. Since few studies on PCS prediction, this section reviews the spatio-temporal relationship modeling methods in non-prediction situations. The concepts involved can provide ideas for future prediction research. Three types: RNN-based methods, CNN-based methods, and methods based on other networks are summarized in the following contents. In practice, due to the complexity of spatio-temporal features, many methods incorporate multiple network structures.

### A. RNN-based Methods

RNN-based methods abstract the spatio-temporal information into states that can be transferred in the RNN structure and attempt to reduce the point cloud's computational cost by various means.

[14] applies LSTM [28] to the task of object detection based on PCS, which uses a sparse convolutional U-Net to extract the spatial features of each frame, and transfer hidden features together with memory features through LSTM to reflect time association. [14] involves voxelization when processing features, while [15] directly handles the original point cloud data structure, using the set abstraction process in [12] to capture spatial features iteratively. Two-layer LSTM is then used to capture temporal dependencies of PCS. Similar to the recognition problem of [15], [29] proposes PointLSTM to combine past state information from neighboring points with the current feature by a wight-shared LSTM layer. [29] further designs a simplified version called PointLSTM-PSS, supposing that the points in the same scene share the same state to solve the problem of time consumption in large-scale scenes. Some research focuses on semantic segmentation. [16] designs a single frame feature extractor to obtain spatial features from range image recursively, combined with an alignment step. Recursive aggregation operation avoids repeated extraction of information, ensuring efficient feature transmitting in the time dimension. Other models use convolutional networks to process structured point cloud data while still introducing the RNN architecture to learn the temporal relationship between frames [30-33].

The RNN architecture is conducive to understanding time-series data, but it often brings enormous calculations. The above methods place extra emphasis on the use of temporal information.

### B. CNN-based Methods

CNN-based methods structurally represent the point cloud data in the form of range images or voxels and regard the time dimension in the PCS as an extra spatial dimension.

[34] performs 3D convolutions on a 4D tensor, which encoding an occupancy grid of 3D space over time frames. While 3D-CNN can extract spatial and temporal features efficiently, it will lead to the missing of information due to its sparse connectivity property. On this basis, [35] proposes a 4D convolutional network called MinkowskiNet, which uses generalized sparse convolutions to process high-dimensional sparse data and cooperates with innovative hybrid kernels to reduce the computation cost when extracting spatio-temporal features. Besides, [35] introduces conditional random fields to enforce spatio-temporal consistency. Instead of using RNN architecture, these methods [34-36] treat time as an additional spatial dimension in representation and use CNN to extract spatio-temporal features simultaneously. For 4D data types, [37] introduces the cross-frame global attention module and cross-frame local interpolation module to associate spatio-temporal frames and highlight the current features. However, this method can only process two consecutive scenes. [38] also introduces an attention mechanism to embed disordered spatial point features into anchors and uses 3D-CNN to extract spatiotemporal features.

Compared with the RNN-based methods, CNN-based methods focus more on the use of spatial dimensions. While CNN is efficient and straightforward, it has three limitations: structured representation compresses part of the spatial and temporal information; challenging to adjust PCS's length; repeated extraction of features in the hierarchical learning process.

### C. Methods Based on Other Networks

Some studies look for novel ways of modeling spatio-temporal features. [39] applies Siamese network for temporal loss calculation, trying to infer temporally coherent features between point cloud frames in generation problems. [33] uses graph neural network (GNN) to construct spatial connections between neighborhoods, combining an expanding receptive field to extract spatial features of different scales. [40] learns spatial and temporal features in a joint framework to realize the prediction of autonomous vehicles' trajectory.

## IV. LOW-LEVEL POINT CLOUD SEQUENCE PREDICTION IN INTERACTIVE SCENARIOS

In contrast with the various environment understanding tasks in Section III, the prediction of PCS needs to restore the hidden features to coordinate space after encoding the spatio-temporal features to truly reflect point-wise movement. According to the output, we divide the low-level PCS prediction into the following two types: Scene Flow Estimation (SFE) and Point Cloud Location Prediction (PCLP), as illustrated in Figure 1. We mainly focus on the methods that can be applied to large-scale interactive scenes of autonomous driving. These methods usually have strong feature extraction ability and can handle more scanning points. The difference and relation between SFE and PCLP will also be discussed.

### A. Scene Flow Estimation (SFE)

Scene flow is the three-dimensional motion field that represents points movement [41]. Scene flow estimation aims to obtain the flow vector of each point at the previous frame by capturing the spatio-temporal feature between two consecutive scenes, which can reflect each point's 3D motion. Some studies explore scene flow through non-learning methods [42, 43]; this section only focuses on learning-based methods that do not require assumptions.

#### a) One-frame to One-frame

[44] proposes FlowNet3D, which extracts spatial features from two consecutive frames using PointNet++ architecture and aggregates the geometric similarities and spatial relations of points through the proposed flow embedding layer. A novel set upconv layer is used to upsampling the flow embeddings to obtain the flow vector of each original point. FlowNet3D tackles the SFE problem on point cloud directly, but the obtained motion vectors often deviate from ground truth

direction. On top of [44], [45] introduces point-to-plane loss to adapt to non-rigid body motion scenarios and considered cosine distance loss to make the predicted direction coincide with the ground truth. [46] uses nearest neighbor loss and cycle consistency loss to minimize the prediction error in [44]. Besides, [46] adopted a self-supervised training approach which allows the method to be trained on large unlabeled autonomous driving datasets.

In the interactive point cloud scenario, the self-supervised learning mode is worthy of attention, since scene flow annotation is hard to acquire in real data [47]. Most methods choose to train the model in a large-scale synthetic dataset and verify it in a large-scale real-scene dataset [44, 45, 48-51]. However, these methods fail to learn the spatio-temporal features in the interactive scenarios at the training level, limiting their generalization ability. In [47], the scene flow is obtained in a coarse-to-fine fashion, which can reflect the large motion to a certain extent. In particular, a loss function combining Chamber distance, Smoothness constraint, and Laplacian regularization is proposed to realize self-supervised training. [52] segments the point cloud scene into two parts, the foreground (moving objects) and the background (static objects), and integrated the flow into a high-level abstract motion to achieve weakly supervised training.

[53] divides the point cloud scene into voxel using VoxelNet [54], learning features by 3D-CNN to perform target detection and voxel-wise rigid SFE. Unlike the feature extraction method in [11], [48] uses BCL [55] to process high-dimensional sparse data in a permutohedral lattice network, which can learn PCS features at different scales to obtain scene flow in large scenes. These two methods transform the irregular point cloud structure into regular input, resulting in the loss of information.

The above methods only focus on the motion correlation between two consecutive point cloud scenes without taking advantage of the rich historical information in PCS.

### b) Multi-frame to One-frame

Since PCS contains many point cloud scenes, it can better characterize the changes in the motion of each point during the continuous transmission of spatio-temporal features. Recognizing the importance of time transfer in timing problem, some studies estimate the scene flow based on PCS.

[51] proposes MeteorNet to learn from multiple point cloud frames. MeteorNet uses direct grouping or chained-flow grouping to delimit the neighbor points, aggregate the temporal and spatial neighborhood information of multiple scenes to reflect the motion, and finally upsampling through feature propagation to get motion vectors. This method can be used for various tasks such as point cloud classification, segmentation, and SFE. The capture of long-sequence information enables MeteorNet to characterize the development of motion in the time dimension. [56] introduces TPointNet++ architecture to learn global and local PCS features and encodes the time dimension separately in the hidden layer, making the output result more continuous and more robust to sampling. [57] expresses PCS as bird's eye view images. The spatio-temporal features of image sequences are extracted hierarchically, and the prediction of flow vectors are smoother in time and space by a spatial and temporal continuous loss.

The above methods encode the spatio-temporal features of multiple point cloud scenes to better express temporal dynamics. The output scene flow can be directly applied to the PCLP task described in the next part.

### B. Point Cloud Location Prediction (PCLP)

Scene flow estimated from PCS can generate new coordinate predictions to reflect the position change of the point cloud at a future time [58]. In the field of autonomous driving, video prediction based on RGB images and depth images is widely used [59, 60]. We only review methods based on point cloud since it has the complete information of space. Some researchers make PCLP on a single object [61, 62]. This type of method needs to contact historical spatio-temporal features to make a reasonable judgment of motion, which is different from the motion segmentation and generation from a single point cloud scene [63, 64]. Compared with the single object PCLP problem, the spatial and temporal relationships in interactive scenes are more complex with a large number of points. We generally require $M > 1, N \geq 1$ in PCLP problem.

[23] employs RNN directly on the point cloud. The proposed PointRNN concatenates current features with past states from neighbor points according to their coordinates and then uses a pooling layer to update current states. This method uses sampling & grouping operation in [12] to reduce the amount of calculation in predicting part. However, [23] lacks the local spatial features in each frame. In contrast, [24] uses content features to characterize the spatial relationship while using motion features to capture the temporal relationship between frames. States are updated by the proposed MotionRNN using "encoder-decoder" architecture. The same architecture is also used in [25], which introduces the DConv operator to extract spatial information from grid-structural point cloud data. The encoded features are fed into the LSTM directly, combined with an attention mechanism. The above methods imply a one-to-one correspondence between points in PCS. Usually, a fixed number of points are randomly selected for training. To process more points, [22] first use PointNet to extract spatial features and then aggregate each point's feature through a combination of Multi-Layer Perceptron (MLP) and max-pooling operator. Such operation gives the model the ability to process large-scale scenes rather than just focusing on some sample points. Based on the above encoding process, [22] uses LSTM to transfer spatio-temporal features. However, this method lacks the extraction of local spatial features.

### C. Difference and Relation between SFE and PCLP

In this paper, SFE and PCLP are regarded as two forms of point cloud prediction. We briefly summarize their difference and relation as follows:

#### a) Difference

SFE describes the motion trend of each point in the past moment, while PCLP obtains the predicted result at the future moment. SFE usually captures the movement changes between two scenes, while PCLP focuses on multiple scenes. SFE generally uses a supervised training approach, while PCLP usually uses a self-supervised learning method.

#### b) Relation

Both SFE and PCLP are point-wise level predictions; the future position of points can be obtained through motion vector; some SFE methods began to learn spatio-temporal features from multiple observations.

## V. Future Outlook and Opening Issues

This section summarizes many opening issues in the sequential point-cloud prediction area and looks forward to the future.

### A. Extraction of Spatio-temporal Features

Existing methods usually use RNN and CNN to model spatio-temporal features. However, CNN-based methods do not consider the time dimension separately, and it is difficult to deal with the change in the number of PCS frames. RNN-based methods suffer from a large amount of calculation, and the information transfer between RNN layers is usually insufficient. The proper use of spatio-temporal information is the key to low-level point cloud prediction. Future research should weigh the impact of spatial and temporal features and focus on more extended time series. Moreover, neural networks with other special characteristics, such as GNN, can be applied to related problems.

### B. Global Prediction of Interactive Scenario

Many methods at this stage cannot handle the entire interactive scene well. To reduce the computational cost, related methods usually sample a fixed number of points in the scene or repeatedly reduce dimensionality and compress features. Such operations have caused the loss of interactive information. In the future, it is necessary to extract features in the large point cloud scenarios at a reasonable computational cost.

### C. Hierarchical Understanding of Environment

The movement of participants in the traffic environment has a vital influence on the decision-making of autonomous vehicles. Therefore, we can focus more on dynamic participants in the point cloud scene under the premise of considering background factors. A hierarchical understanding of the interactive point cloud scenario can make the model more focus on the primary research object while weakening the influence of other environmental factors.

### D. Multiple Learning Method

Most PCS prediction methods adopt self-supervised learning mode. However, scene flow estimation is generally based on supervised learning, which brings higher cost and limited generalization ability in real interactive scenarios. Future research should continue to explore the self-supervised learning approach in the SFE problem and try to combine learning methods such as transfer learning and attention mechanisms to improve learning efficiency. Other potential directions, such as heterogeneous prediction and out-of-distribution prediction may also be included.

## VI. Conclusion

The rapid development of autonomous driving technology and the popularity of LiDAR sensors have made point clouds more used in environmental understanding tasks. This article systematically reviews point cloud prediction methods, emphasizing the importance of their generalization ability in interactive scenarios. From spatio-temporal features to point cloud predictions, we summarize the relevant prediction methods from two levels while highlighting the correlation between the methods. At the end of this article, we put forward the outlooks of future research. Sequential point cloud prediction can provide dynamic information of the surrounding environment, which can help autonomous driving vehicles understand the changes of the scene and make human-like decisions. The obtained point cloud scene in the future can help autonomous vehicles to avoid obstacles and make driving decisions through the combination of semantic segmentation, target recognition and other methods.